\documentclass[runningheads]{llncs}
\usepackage[T1]{fontenc}
\usepackage{graphicx}
\usepackage{enumitem}
\setlist{nosep}
\usepackage{multirow}%
\usepackage{amsmath,amssymb,amsfonts}%
\usepackage{mathrsfs}%
\usepackage[title]{appendix}%
\usepackage{xcolor}%
\usepackage{textcomp}%
\usepackage{manyfoot}%
\usepackage{booktabs}%
\usepackage{algorithm}%
\usepackage{algorithmicx}%
\usepackage{algpseudocode}%
\usepackage{listings}%
\usepackage{enumitem}
\setlist{nosep}

\usepackage{caption}
\usepackage[hidelinks]{hyperref}

\usepackage{orcidlink}
\usepackage{cite}

\newcommand{\repthanks}[1]{\textsuperscript{\ref{#1}}}
\makeatletter
\patchcmd{\maketitle}
{\def\thanks}
{\let\repthanks\repthanksunskip\def\thanks}
{}{}
\patchcmd{\@maketitle}
{\def\thanks}
{\let\repthanks\@gobble\def\thanks}
{}{}
\newcommand\repthanksunskip[1]{\unskip{}}
\makeatother

\begin{document}

\title{Hybrid Compression: Integrating Pruning and Quantization for Optimized Neural Networks}

\titlerunning{Hybrid Compression for Optimized Neural Networks}

\author{
Minh-Loi Nguyen \thanks{These authors contributed equally to this research. \protect\label{X}}\inst{1,2}\orcidlink{0009-0003-2630-3325}
\and Long-Bao Nguyen \repthanks{X}\inst{1,2}\orcidlink{0009-0003-6311-745X}
\and Van-Hieu Huynh \repthanks{X}\inst{1,2}\orcidlink{0009-0002-7790-4414}
\and Minh-Triet Tran \inst{1,2}\orcidlink{0000-0003-3046-3041}
\and Trung-Nghia Le \thanks{Corresponding author.} \inst{1,2}\orcidlink{0000-0002-7363-2610}
}

\authorrunning{M.-L. Nguyen et al.}

\institute{
University of Science, Ho Chi Minh city, Vietnam \and Vietnam National University, Ho Chi Minh city, Vietnam\\
\email{\{22120189,22120025,22120105\}@student.hcmus.edu.vn}, 
\email{\{tmtriet,ltnghia\}@fit.hcmus.edu.vn}
}

\maketitle

\begin{abstract}
Deep neural networks have witnessed remarkable advancements in recent years and have become integral to various applications. However, alongside these developments, training and deployment of neural network models on embedding and edge devices face significant challenges due to limited memory and computational resources. These problems can be addressed with deep neural network compression, which involves a trade-off between model size and performance. In this paper, we propose a novel method for model compression through two phases. First, we utilize model compression techniques, such as pruning and quantization, to significantly reduce the model size. Then, we use Mixture of Experts to route the previously compressed models to enhance performance while maintaining a balance in inference efficiency. MoEs consist of multiple expert models (i.e., compressed models) that are moderately sized and deliver stable performance. Experimental results on several benchmark datasets show that our method successfully compresses CNN models which achieves substantial reductions in FLOPs and parameters with a negligible accuracy drop.


\end{abstract}

\section{Introduction}

\begin{figure}[!t]
\centering
\includegraphics[width=0.8\textwidth]{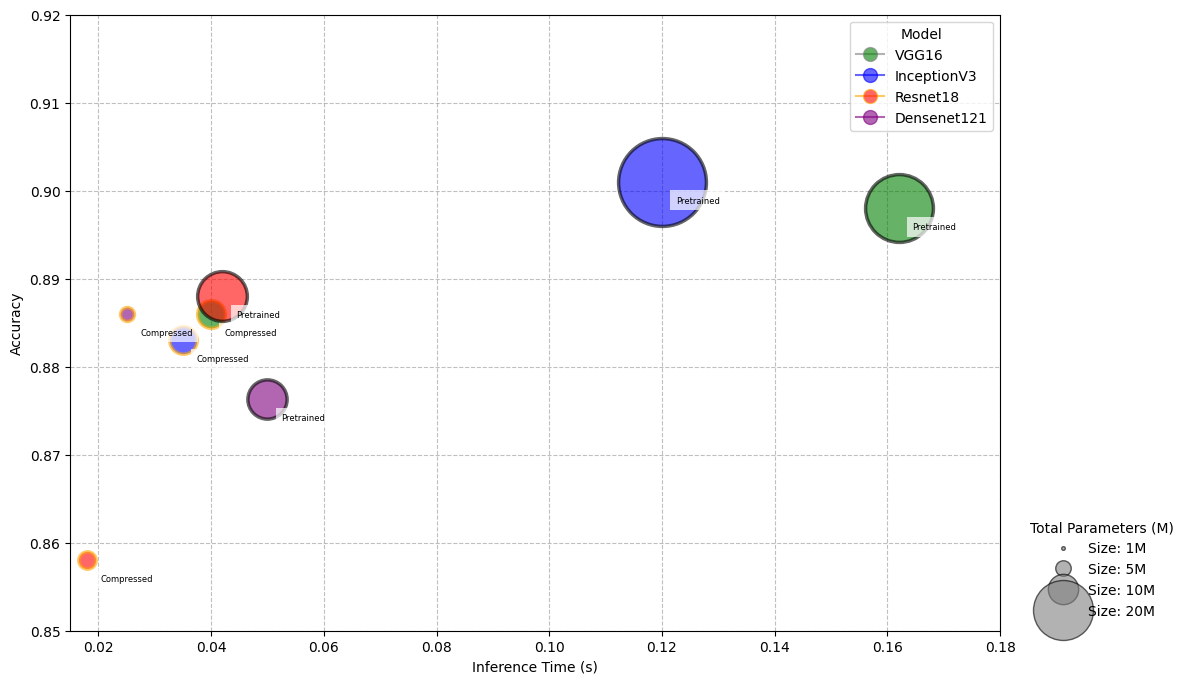}
\caption{Comparison of CNN model performance on CIFAR-10 dataset. Our proposed method for model pruning and quantization achieves significant reduction in FLOPs, total parameters, and inference time, while maintaining accuracy with minimal dropout compared to the original models.}
\label{figure:teaser@}
\end{figure}


Over the past decade, the explosion of data and computational power has driven significant advancements in deep neural networks (DNNs) \cite{c10}. As a result, many new network architectures have emerged, each more complex and demanding more resources than its predecessors \cite{c11}. For instance, the first Convolutional Neural Network (CNN) model was proposed in 1998 with fewer than 1 million parameters \cite{c15}, while OpenAI's GPT-3 model in 2020 comprised up to 175 billion parameters, requiring hundreds of gigabytes of memory for storage and thousands of teraflops for training. The rapid development of these large-scale models has introduced significant challenges and limitations. Deploying DNN models in real-world scenarios, such as mobile applications and Internet of Things (IoT) devices, often becomes impractical due to constrained memory and computing resources \cite{c11}.

To address these challenges, the field of model compression has gained considerable attention \cite{c13}. Model compression techniques aim to reduce the size and computational requirements of DNNs without significantly compromising their performance. Among these techniques, deep compression has emerged as a robust approach, with methods such as pruning, quantization, and the Mixture of Experts (MoE) achieving substantial reductions in model size and computational cost \cite{b10}. Pruning methods \cite{b6} are designed to remove less important connections in neural network layers based on various evaluation criteria. Instead of using high-precision floating-point numbers, quantization methods \cite{b12, b10} reduce the precision of parameters by representing them with fewer bits. MoE \cite{c1, c2} dynamically selects a subset of parameters (or experts) for each input, optimizing resource use by activating only the network parts relevant to each task, enabling efficient scaling.

In this paper, we introduce a novel multi-stage method to develop a cost-efficient CNN-based model. Our approach focuses on optimizing both the complexity and the computational efficiency of the model through a series of targeted stages. In the first stage, we employ well-established hard compression techniques such as pruning and quantization to significantly reduce the model's complexity, including the number of parameters and the overall inference cost, making it more feasible for deployment in resource-constrained environments. In addition, we leverage the Neural Network Intelligence (NNI) \cite{b11} framework to implement and automate our pruning and quantization techniques. The second stage involves the MoEs paradigm to enhances the model's adaptability and efficiency by allocating previous compressed model to specialize for each input. The specialization helps enhance the performance and stability of compressed models, which might drop due to pruning and quantization, while still leveraging the low resource consumption and computational cost of these compressed models. Experimental results on CIFAR-10 \cite{b13} and BloodMNIST \cite{c17} datasets show that our method successfully achieved a 10x-11x reduction in FLOPs and a 10.5x reduction in parameters, with a negligible accuracy drop on the image classification task (See Fig.~\ref{figure:teaser@}).

In summary, our contributions are as follows:
\begin{itemize}
\item We introduce a novel method that combines pruning, quantization, and the Mixture of Experts (MoE) paradigm, demonstrating how this fusion brings superior effectiveness and provides detailed insights into the trade-offs between model size, computational efficiency, and accuracy.
\item We investigate our method on different CNN models, providing practical insights for implementing compression model techniques.
\end{itemize}


\section{Related Work}

Pruning is a widely used technique for compressing neural networks by removing redundant weights and connections. These methods identify unimportant elements in the model, such as weights and neural connections, and prune them by setting their values to zero, ensuring they do not participate in the backpropagation process. Hassibi et al. \cite{b25} introduced an early pruning method that uses the inverse Hessian matrix to identify and remove redundant weights, while updating the remaining ones with second-order information. More recently, various pruning techniques have emerged, including magnitude-based weight pruning \cite{b6}, which gradually eliminates small magnitude weights to achieve network sparsity. In CNN models, pruning is typically categorized into two approaches: weight pruning \cite{b28}, which removes individual redundant weights, and filter pruning \cite{b30}, which eliminates entire convolutional filters with minimal impact on performance.

Quantization is a popular technique for compressing neural networks by lowering parameter precision, reducing memory usage and computational costs. Binarized neural networks Quantization method \cite{b34} trains networks with binary weights and activations but still accumulates gradients in 32-bit precision, highlighting the need for high precision during training. DoReFa-Net Quantizer \cite{b35} reduces gradient precision by quantizing them into low-bitwidth floating-point numbers. Quantization methods generally fall into two categories: Quantization-Aware Training (QAT) \cite{b10}, where models are retrained with quantized weights and activations, and Post-Training Quantization (PTQ) \cite{b12}, which quantizes a pretrained model without retraining. This paper focuses on the QAT approach for CNN model quantization.

As network architectures evolve, combining them for improved performance is gaining traction, with the Mixture of Experts (MoE) method \cite{c3} being a prominent approach. Unlike traditional ensemble methods like bagging or boosting, where all models contribute equally, MoE uses a gating network to select specific experts for different problem areas, enhancing performance with fewer computational resources \cite{c16, c3}. Classic MoE models \cite{c1} consist of multiple expert models and a gating network, while Eigen et al. \cite{c2} extended this by incorporating MoE subcomponents with their own gating.

In natural language processing, Shazeer et al. \cite{c3} introduced Sparse Mixture of Experts (SMoE), replacing dense layers in Transformers with MoE layers. For example, Mixtral 8x7B \cite{c16}, an SMoE-based model, performs comparably to Llama 2 70B and GPT-3.5. In computer vision, particularly CNNs, MoE has also shown success, as seen in DeepMoE \cite{c9}, where MoE layers replace traditional convolutional layers, with a multi-headed gating network optimizing channel selection.

\section{Methodology}



\begin{figure}[!t]
\centering
\includegraphics[width=\textwidth]{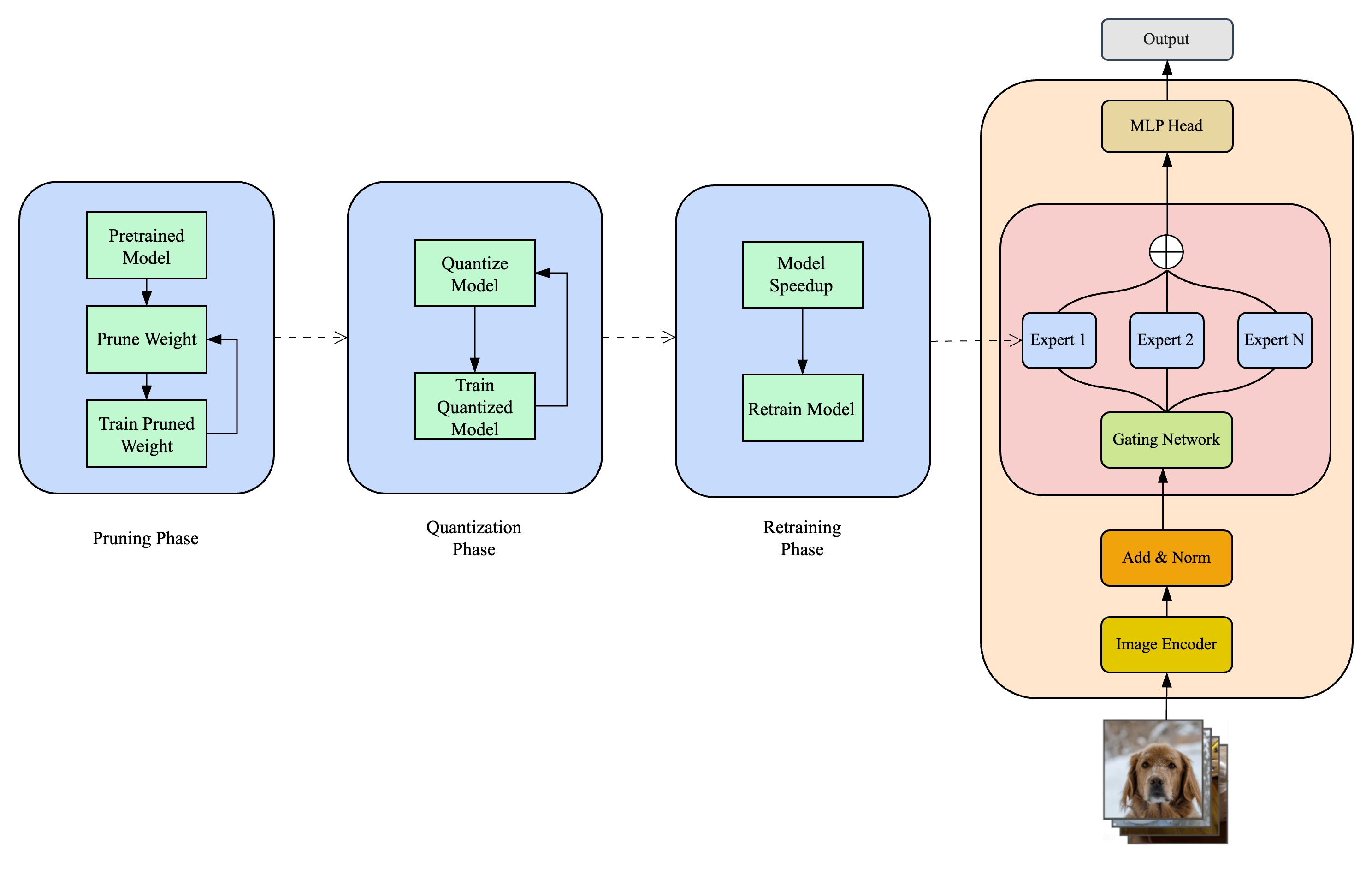}
\caption{Overview of our proposed three-phase deep compression method, including pruning, quantization, and the Mixture of Experts (MoE) paradigm.}
\label{figure:1}
\end{figure}

\subsection{Overview}


We present an in-depth exploration of the methodologies employed in our research, aim to compress CNN models for deployment on resource-constrained hardware, such as mobile and edge devices. As seen in Fig. \ref{figure:1}, our proposed method consists of three phases: pruning, quantization, and integrating the MoE paradigm; each is designed to progressively compress CNN, enhance efficiency, and ensure the robustness of the model.

\begin{itemize}


    \item \textbf{Pruning Phase:} by applying iterative pruning algorithms, such as Automated Gradual Pruning (AGP) \cite{b3}, we systematically remove less important connections. The outcome is a model with reduced computational requirements, mitigating the risk of layer collapse and preserving essential network structures.

    \item \textbf{Quantization Phase:}  Following pruning, we apply the QAT technique \cite{b10} to reduce further both the model's size and its computational resource requirements. This involves representing the model's weight with lower precision, typically using fewer bits. 

    \item \textbf{Mixture of Experts (MoE) Integration: } finaly, we introduce the MoE paradigm to enhance the performance and efficiency of the compressed model. The MoE approach dynamically selects a subset of experts for each input, optimizing resource utilization and maintaining high performance. 
\end{itemize}

\subsection{Pruning} Network pruning, crucial for reducing model size and bandwidth needs, removes unnecessary neural connections \cite{b14}. We use Magnitude Pruning \cite{b26} to eliminate redundancies effectively.

\textbf{Automatic Gradual Pruning (AGP)}: Large neural networks often contain redundant parameters, as shown in prior work \cite{b6, b14}. Frankle et al. \cite{b4} proposed that large networks have a subnetwork capable of matching the original’s performance, while Arora et al. \cite{b7} demonstrated that over-parameterized models retain generalization when compressed. Thus, we apply sparsity levels between 50\% and 70\%, but high sparsity can lead to layer collapse, where entire layers are removed under one-shot pruning \cite{b5}.

To prevent this, we adopt the Automatic Gradual Pruning (AGP) algorithm \cite{b3}, an iterative framework that removes redundant connections gradually with a cubic function, adjusting sparsity at each step. At each training step, magnitude pruning \cite{b26} globally prunes CNN connections, avoiding model collapse while allowing recovery through concurrent training.

\begin{equation}
s_t = s_f + (s_i - s_f)\left(1 - \dfrac{t - t_0}{n \Delta t}\right)^3, 
\end{equation}
where $n$ is the number of pruning steps, $s_i$ is the initial sparsity level, $s_f$ is the final sparsity level, $s_t$ is the sparsity level at pruning step t, which is updated for every $\Delta t$ steps, and $t$ is current pruning step whereas $t \in \{t_0\, ,t_0 + \Delta t\,, \ldots\, , t_0 + n\Delta t\}$.

The AGP’s formula indicates that the sparsity level increases gradually during the initial stages, leading to fewer redundant connections being removed. This gradual increase allows the model to adapt and learn from the pruned information, maintaining a balance in the importance scores across layers \cite{b5}. This balance prevents layer-collapse, as the importance scores among layers remain equivalent. In the final stage, AGP pruning imposes a high sparsity to achieve the configured target level.

We also consider that pruning can potentially disrupt the neural network's structure, resulting in a substantial decrease in accuracy. This challenge can be addressed by retraining the model, which incurs additional costs \cite{b6, b14}. In our pipeline, retraining is performed after the model undergoes the Speedup technical, which optimizes the model for faster execution. Additionally, during model construction, we observed the weight distribution of the classifier layer, responsible for generating the logit vector for classification output. Therefore, in our configuration, we set a target sparsity level for the entire CNN backbone and a lower sparsity level for the classifier layer.

\textbf{Model Speedup}: In the pruning process, a binary mask layer is used to represent retained connections, assigning a value of 1 to kept connections and 0 to reduced ones. Consequently, during the forward and backward passes of the model, this binary mask matrix is multiplied with the corresponding weights. Obviously, this pruning method does not significantly enhance model inference and training speed.

To mitigate this limitation, we utilize the Model Speedup method  \cite{b17}, which involves removing the feature maps that were previously pruned in the CNN layer and retaining the weights to preserve the layer's output. As a result, the model achieves a smaller weight set than the original. This optimization can lead to a latency reduction by a factor of 2 compared to the original model, albeit with a slight trade-off in accuracy.

\subsection{Quantization}

Quantization reduces model size and speeds up inference by converting weights or activations from high-precision floating points to lower bit-widths, like 8-bit integers, with minimal accuracy loss \cite{b10, b12}.

In this paper, we implement QAT \cite{b10}  to maintain high accuracy post-quantization. QAT \cite{b10} simulates the effects of quantization during training, allowing the model to learn and adjust to the reduced precision, which minimizes the accuracy degradation typically observed in post-training quantization \cite{b6,b12}. 

\textbf{Quantization Process}: The quantization process involves mapping the floating-point values to discrete integer values. This is achieved through two main steps: scaling and rounding. The transformation for a value $x$ in floating-point format to an 8-bit integer is given by:

\begin{equation}
\tilde{x} = \left\lfloor \frac{x - \min(x)}{\Delta} \right\rfloor,
\end{equation}
where $\min(x)$ is the minimum value in the range of $x$, and $\Delta$ is the quantization step size defined as:

\begin{equation}
\Delta = \frac{\max(x) - \min(x)}{2^n - 1},
\end{equation}
where $n$ is the bit-width (e.g., 8 for 8-bit quantization) \cite{b10}. The quantized value $\tilde{x}$ is then converted back to a floating-point format during inference using $x_q = \tilde{x} \cdot \Delta + \min(x),$ ensuring the model operates within the quantized value range during execution.


\textbf{Fake Quantization in QAT. }
In QAT, we introduce \textit{fake quantization} operations during training to mimic the effects of quantization on weights and activations. This operation is expressed as:

\begin{equation}
\tilde{x}_{\text{fake}} = \Delta \cdot \left\lfloor \frac{x}{\Delta} \right\rceil,
\end{equation}
where $\left\lfloor \cdot \right\rceil$ denotes the rounding operation to the nearest integer \cite{b6,b38}. This transformation ensures that the model learns to adapt to the quantized weights and activations during training.

\textbf{Straight-Through Estimator (STE)}: To facilitate backpropagation through quantized nodes, we use the Straight-Through Estimator (STE) \cite{b6,b39}. The STE approximates the gradient of the quantization function with respect to the input as:

\begin{equation}
\frac{\partial \tilde{x}_{\text{fake}}}{\partial x} \approx 1 \text{ if } |x| \leq 1, \text{ else } 0.
\end{equation}

This approximation allows gradients to flow through the quantization operation, enabling effective training of the quantized model.

\textbf{Scale Factor and Zero Point Initialization}: To ensure the quantization parameters are suitable for the data distribution, we initialize the scale factor $\Delta$ and zero point $z$ using calibration data. The scale factor is computed as:

\begin{equation}
\Delta = \frac{\max(x_{\text{calib}}) - \min(x_{\text{calib}})}{2^n - 1}.
\end{equation}

The zero point is calculated to align the quantized range with the original range $z = \left\lfloor -\frac{\min(x_{\text{calib}})}{\Delta} \right\rceil.$


\textbf{Loss Function}:
Our implemeted loss function is as follow:

\begin{equation}
\mathcal{L}_{\text{total}} = \mathcal{L}_{\text{task}} + \lambda \cdot \mathcal{L}_{\text{quant}},
\end{equation}
where $\mathcal{L}_{\text{task}}$ is the task-specific loss (e.g., cross-entropy for classification), $\mathcal{L}_{\text{quant}}$ is the quantization loss, and $\lambda$ is a balancing hyperparameter \cite{b6,b39}.

\subsection{Retraining}

Retraining is an essential step in the model compression pipeline to recover the accuracy lost during pruning and quantization. After applying pruning and quantization, the model may suffer from reduced performance due to significant structural and precision changes. Retraining enables the model to regain performance by re-optimizing its weights within the new compressed architecture. This improves accuracy by allowing the model to adapt to altered parameters and mitigate errors introduced during quantization \cite{b6, b14}.

\subsection{Mixuture of Experts (MoE)}
The MoE framework leverages a set of neural network blocks, known as experts $E_1, E_2, \ldots, E_n$, along with a gating network $G$ that determines which experts should be activated for a given input (See Fig.~\ref{figure:1}). The gating network outputs an n-dimensional distribution vector, which directs the input to specific experts based on their relevance to the task at hand. While traditional MoE implementations typically select feed-forward networks as experts, our approach integrate pre-compressed models into the MoE architecture. This design choice allows us to leverage the computational efficiency of compressed models while benefiting from the diverse expertise each model brings, thus enhancing overall performance.

In this paper, we selected a variety of models to undergo compression and be integrated into the MoE framework. This strategy not only reduces the computational overhead associated with using full-sized models but also provides a rich diversity of learned representations that can be exploited by the gating network. However, a significant challenge in this approach is the potential imbalance among experts, where certain experts may dominate the inference process, leading to what is known as expert collapse. This phenomenon can diminish the diversity of the activated experts, ultimately affecting the model’s performance and generalization capability.

To address this issue, we implemented the Top-k Noisy Gating algorithm \cite{c3} as our routing strategy. This algorithm introduces tunable Gaussian noise to the logits before applying the softmax function, ensuring that the gating decisions are not deterministic but rather probabilistic, allowing for a more balanced selection of experts. By selecting the top-k experts based on their noisy scores and setting the rest to $\inf$, we promote a more balanced activation of experts, reducing the risk of expert collapse and ensuring that a diverse set of experts contribute to the inference process:

\begin{equation}
    G(x)=\operatorname{Softmax}(\operatorname{TopK}(H(x), k)),
\end{equation}
\begin{equation}
H(x)_i=\left(x \cdot W_g\right)_i+ \epsilon \cdot \operatorname{Softplus}\left(\left(x \cdot W_{\text {noise }}\right)_i\right),
\end{equation}
\begin{equation}
    \text { TopK }(v, k)_i= \begin{cases}\, v_i & \text { if } v_i \text { is in the top } k \text { elements of } v \\
-\infty & \text { otherwise}\end{cases},
\end{equation}
where $\epsilon$ denotes the standard normal random noise while $W_g$ and $W_{\text {noise }}$  represent weights for the gating mechanism and noise, respectively.

\section{Experimental Results}
\subsection{Implementation Details}

We conducted our experiments using PyTorch to redefine common CNN backbones. For each CNN model, we sequentially perform the steps of defining, pretraining, pruning, quantization, and finetuning the compressed model. 
Additionally, we leverage the Neural Network Intelligence (NNI) \cite{b11} framework to implement and automate our pruning and quantization techniques.

\subsection{Datasets}

We utilized a diverse set of datasets to validate the robustness and generalizability of our proposed model compression techniques across different domains. 
\textbf{CIFAR-10 \cite{b13}} consists of 60,000 color images with resolution $32 \times 32$ in 10 different classes, with 6,000 images per class. 
\textbf{BloodMNIST \cite{c17}} includes 17,092 images of normal cells captured using the CellaVision DM96 analyzer at the Hospital Clinic of Barcelona and grouped into 8 categories. 


\begin{table}[t!]
\centering
\caption{Experimental results on CNN models.}
\label{tab:Table_1}
\resizebox{\textwidth}{!}{%
\begin{tabular}{|l|c|cc|cc|cc|}
\hline
\multirow{2}{*}{\textbf{Method}} & \multirow{2}{*}{\textbf{Total Parameters}} & \multicolumn{2}{c|}{\textbf{FLOPs ↓}} & \multicolumn{2}{c|}{\textbf{Inference speed (s)}} & \multicolumn{2}{c|}{\textbf{Accuracy (\%)}} \\
 & & \textbf{CIFAR10} & \textbf{BloodMNIST} & \textbf{CIFAR10} & \textbf{BloodMNIST} & \textbf{CIFAR10} & \textbf{BloodMNIST} \\
\hline
VGG16 (Original) & 21.15M & -- & -- & 0.162 &  0.0055 & 89.8 & 94.5 \\
VGG16 (Ours) & 3.3M & x10.71 & x10.76 & 0.041 & 0.0045 & 88.6  & 90.0 \\
\hline
Resnet18 (Original) & 11.2M & -- & -- & 0.042 &  0.007 & 88.8 & 96.1 \\
Resnet18 (Ours) & 1.01M & x10.6 & x11.0 & 0.018 & 0.0059 & 85.8 & 95.6 \\
\hline
InceptionV3 (Original) & 35.37M & -- & -- &  0.068 & 0.045 & 86.6 & 93.9 \\
InceptionV3 (Ours) & 3.23M & x10.86 & x10.81 & 0.037 & 0.032 & 90.3 & 90.7 \\
\hline
Densenet121 (Original) & 6.87M & -- & -- & 0.052 &  0.042 & 87.6 & 92.5 \\
Densenet121 (Ours) & 0.648M & x10.51 & x10.56 & 0.037 & 0.036 & 88.6 & 95.8 \\
\hline
MOE (Ours) & & & & & & 92.1 & 96.9 \\ 
\hline
\end{tabular}
}
\end{table}


\subsection{Results}


Table \ref{tab:Table_1} shows our extensive experiments on CIFAR-10 and BloodMNIST datasets. The experimental results indicate that our proposed compression method yields significant FLOPs and inference time reduction across multiple models while maintaining the high performance.

Our compressed and finetuned models usually perform well compared to the original models. The accuracy of VGG16 and Resnet18 slightly drops from 1.2\% to 4.5\% and from 0.5\% to 3.0\%, respectively, indicating that our compression method does not significantly harm the model's ability to make accurate predictions. In fact, in some cases like InceptionV3 and Densenet121, the accuracy even improves from 1.8\% to 3.7\% and from 1.0\% to 1.1\%. These results demonstrate that our compression method preserves, and in some cases enhances, the performance even with substantial reductions in model size, making it highly effective for tasks on resource-limited platforms.


The reduction in FLOPs is a significant highlight of our compression approach. For all CNN models, the reduction in FLOPs consistently achieves a factor of from 10.5 to 11.0, confirming the effectiveness of our compression techniques in enhancing computational efficiency. On the other hand, inference speed improvements are evident in the results, with compressed models exhibiting faster prediction times than their original counterparts. These findings underscore the practical advantages of compression for real-time applications, where quicker inference can greatly improve the user experience.



To address accuracy dropout, we decided to integrate our compressed and finetuned models with the MoE algorithm. The results in Table \ref{tab:Table_1} shows that our implemented MoE method achieved an accuracy of 92.1\% on CIFAR-10 and 96.9\% on the BloodMNIST dataset, successfully increased accuracy levels comparable to the original models before compression while leveraging the computational efficiency of the compressed models.


\section{Conclusion}

We have presented a novel deep learning model compression method that combines pruning, quantization, and the Mixture of Experts (MoE) paradigm. Our approach significantly reduces model size and computational requirements without compromising accuracy. 
Experimental results demonstrated the potential of our method for deploying sophisticated deep learning models on resource-constrained devices. Our approach enables the use of complex neural networks in mobile and edge applications where computational resources and energy efficiency are critical constraints.

\section*{Acknowledgment}
This research is supported by research funding from Faculty of Information Technology, University of Science, Vietnam National University - Ho Chi Minh City.

\bibliographystyle{splncs04}
\bibliography{sn-bibliography}


\end{document}